# Efficient Inference in Large Discrete Domains

**Rita Sharma**
Department of Computer Science
University of British Columbia
Vancouver, BC V6T 1Z4
rsharma@cs.ubc.ca

**David Poole**
Department of Computer Science
University of British Columbia
Vancouver, BC V6T 1Z4
poole@cs.ubc.ca

## Abstract

In this paper we examine the problem of inference in Bayesian Networks with discrete random variables that have very large or even unbounded domains. For example, in a domain where we are trying to identify a person, we may have variables that have as domains, the set of all names, the set of all postal codes, or the set of all credit card numbers. We cannot just have big tables of the conditional probabilities, but need compact representations. We provide an inference algorithm, based on variable elimination, for belief networks containing both large domain and normal discrete random variables. We use intensional (i.e., in terms of procedures) and extensional (in terms of listing the elements) representations of conditional probabilities and of the intermediate factors.

## 1 Introduction

Bayesian networks [Pearl, 1988] are popular for representing independencies amongst random variables. They allow compact representation of joint probability distribution, and there are algorithms to exploit the compact representations. Recently, there has been much interest in extending the belief networks by allowing more structured representations of the conditional probability of a variable, given its parents (for example, in terms of causal independence [Zhang and Poole, 1996] or contextual independence [Boutilier, Froedman, Goldszmidt and Koller, 1996]). In all of these approaches, discrete random variables are considered to have a bounded number of values.

Some real-world problems contain random variables with large or even unbounded domains, for example, in natural language processing where outcomes are words drawn from large vocabularies. Here, we could have a random variable whose domain is the set of all words (including those words we have never encountered before). As another example, consider the problem of person identification [Gill, 1997; Bell and Sethi, 2001], which is the problem of comparing a test person's description with each person's description in the database. When comparing two records, we have two hypotheses: both records refer to the same person, and the records refer to different people. In a dependence model, where the two descriptions refer to the same person, random variables such as *actual first name*, *actual last name*, and *actual date of birth* are large variables. The domain of *actual first name* may be the set of all possible first names, which we may never know in full extent because people can make up names. In person identification, we can ask, what is the probability of the actual name of a person given the name that appears in the description of the person, or, what is the probability that the two descriptions refer to the same person.

There has been much work on this problem in the context of natural language processing. For an $N$-gram model, for $M$ words vocabulary, there are $M^N$ N-grams and many of such pairs have negligible probabilities. In language processing these models are represented (stored) using efficient N-gram decoding [Odell, Violative and Woodland, 1995] and hash table [Cohen, 1997]. Unfortunately these approaches do not extend to other domains such as the person identification problem.

We assume that we have a procedural way for generating the prior probabilities of the large variables (perhaps conditioned on other variables). This may include looking up tables. For example, the U.S. Census Bureau[1] publishes a list of all first names, conditioned by gender, together with probabilities that covers 90% of all first names for both males and females. This, together with a method for estimating the probability of a new name, can be used as the basis for $P(FirstName|Sex)$. If we have a database of words and empirical frequencies, we can use this, using, for example, a *Good-Turing* estimate [Good, 1953] to compute $P(word)$. We may also have a model of how postal codes are generated to give a procedure that estimates the proba-bility of a given postal code. While we need to reason with the large variables, we never want to actually enumerate the values during inference.

---

[1] http://www.census.gov/genealogy/names/

The fundamental idea is that in any table, we divide the possible values of an unbounded variable in disjoint subsets (equivalence classes) for which we have the same conditional probability for particular values (or particular subsets) of other random variables. We construct these subsets dynamically during inference using the observed states of other variables, or the partitions of other variables in other functions. These subsets are described either as *extensionally* (by listing the element) or *intensionally* (using a predicate).

The remainder of this paper is organized as follows. We first describe the person identification problem, in brief, which motivates the need for the efficient inference for large discrete domains. We then describe the representation for large conditional probability tables. Next we give the details of the inference algorithm followed by the conclusion.

## 2 Motivating Example: Person Identification

Person identification is used for comparing records in one or more data files, removing duplicates, or in determining if a new record refers to a person already in the database or to a new person. The core sub-problem of person identification is the problem of comparing a test person's description with each other description in the database. Let $X$ and $Y$ be two records to be compared, and $Desc_X$ and $Desc_Y$ denote their corresponding descriptions. There are two hypotheses when we compare the two descriptions $Desc_X$ and $Desc_Y$:

- both records refer to the same person ($X = Y$)
- the records refer to different people ($X \neq Y$)

Let $P_{same}$ be the posterior probability that records $X$ and $Y$ refer to the same person given their descriptions and $P_{diff}$ be the posterior probability that records $X$ and $Y$ refer to different people given their descriptions. That is,

$$P_{same} = P(X = Y | Desc_X, Desc_Y)$$

$$P_{diff} = P(X \neq Y | Desc_X, Desc_Y)$$

The odds, $Odds$, for hypotheses $X = Y$ and $X \neq Y$

$$Odds = \frac{P_{same}}{P_{diff}}$$
$$= \frac{P(Desc_X | Desc_Y \wedge X = Y) P(X = Y)}{P(Desc_X | Desc_Y \wedge X \neq Y) P(X \neq Y)}$$

Traditional methods [Fellegi and Sunter, 1969] treat the attributes as independent given whether the desciptions refer to the same person or not. We have relaxed this assumption to model how the attributes are interdependent. We model the dependence/independence between the attributes for both cases $X = Y$ and $X \neq Y$ using a similarity network representation [Geiger and Heckerman, 1996].

To make this paper readable, we only consider the attributes *first name (Fname)* and *phone number (Phone)*. The real application considers many more attributes.

The simplest Bayesian network of attribute dependence for the case $X \neq Y$ does not contain any large variables, and the inference in the network can be done using a standard Bayesian inference algorithm.

Consider the $X = Y$ case where both records refer to the same person (the numerator of the *Odds* formula). If records $X$ and $Y$ refer to the same person, we expect that the attributes values should be the same for both $X$ and $Y$. However, there may be differences because of attribute errors: typing errors, phonetic errors, nick names, swapping first and last names, change of address, and so forth.

We assume that the attributes are dependent because the data entry person could have been sloppy, and because the person could have moved to a new place of residence between the times that the records were input. To make this paper more readable, we consider the following errors[2]: *copy error*[3] (*ce*), *single digit/letter error*(*sde*), and the lack of any errors, or *no error*(*noerr*).

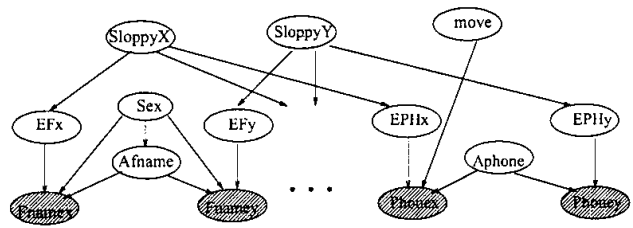

Figure 1: Bayesian Network representation of attribute dependency for case $X = Y$ (shaded nodes are observed)

The dependence between attributes is shown in Figure 1. The unshaded nodes show the hidden variables. The variable *SloppyX* (*SloppyY*) represents whether the person who reported the attribute values of record $X$ ($Y$) was sloppy or not. The variable *Afname* represents the actual first name. The variable *EFx* (*EFy*) represents which error was made in recording the first name for record $X$ ($Y$). The variable *move* represents whether the person moved to a different address between the two records.

---

[2] Although, the real application consider many more errors.
[3] An error where a person copies a correct name, but from the wrong row of a table.

Figure 1 shows the relationship between these variables. The random variables $Fnamex$, $Fnamey$, and $Afname$ have, as domains, all possible first names.

For the probability $P(Afname|Sex)$, we use the first name lists from the U.S. Census Bureau [4]. There are two first name lists with associated probabilities: one for female names, and the other for male names. The probability $P(Afname|Sex = male)$ is computed using the male name file. The probability $P(Afname|Sex = female)$ is computed using the female name file. We need a different mechanism for names that do not appear in these lists. A number of approaches have been proposed to solve this problem [Chen and Goodman, 1998; Good, 1953; Friedman and Singer, 1998]. In our implementation, we just use a very small probability[5] as the estimate of the probability of a new word.

To compute the probability $P(Aphone)$ a model for generating phone numbers can be used. There are rules to generate the valid phone numbers for a city, province, and so forth. We use the simple procedure $P(Aphone)$ is $1/P$, where $P$ is the number of legal phone numbers if $Aphone$ is a legal phone number and is 0 otherwise.

Inference in the Bayesian network shown in Figure 1 is complicated because of the variables with large number of values. We cannot represent $P(Fnamex|Afname \land Sex \land EFx)$ in a tabular form as we do not know all names, and even if we did, the domains of $Afname$ and $Fnamex$ are very large (unbounded). The conditional probability table $P(Afname|Sex)$ is also very large. To represent the large CPTs we need a compact representation.

## 3 Representation

We divide the discrete random variables into two categories **small variables** (small domain size) and **large variables** (large domain size). For **small variables** we treat each value separately (i.e., equivalently partition into single element subsets). For **large variables** we *partition* the values into non-empty disjoint sets (equivalence classes whose union is the domain of the variable). An element of a *partition* is referred to as a *block*.

We use upper case letters to denote random variables (e.g., $X_1$, $X2$, $X$), and the actual value of these variables by the small letters (e.g. $a$, $b$, $x_1$). The domain of a variable $X$, written $dom(X)$, is a set of values. We use the notation $P(X)$ to denote the probability distribution for $X$. We denote sets of variables by boldface upper case letters (e.g. $\mathbf{X}$) and their assignments by the bold lower case letters

(e.g., $\mathbf{x}$).

Each *block* of a *partition* is described either as:

- *intensionally* as a predicate, but we also assume there is a procedure to efficiently compute the predicate, and to count the number of values for which it is true. As a part of the intensional definition, we assume that we have an *if-then-else* stucture, where the condition is a predicate.

- *extensionally* by listing the elements.

The probability is described either as:

- a non negative real number

- intensionally as a function, but we also assume there is a procedure to compute the function.

Let us first consider the representation of the conditional probability table $P(Fnamex|Afname \land Sex \land EFx)$ from BN, shown in Figure 1. We can represent the conditional probability table $P(Fnamex|Afname \land Sex \land EFx)$ by enumerating the following separate cases (i.e., it is an if-then-else structure, where the conditions are on the value of $EFx$):

**case 1:**　$EFx = noerr$

$$P(Fnamex|Afname \land Sex \land EFx = noerr) =$$

$$\begin{cases} 1 & \text{if } equal(Afname, Fnamex) \\ 0 & otherwise \end{cases}$$

where, $equal$ is a predicate to test whether variables $Fnamex$ and $Afname$ have the same value or not. If the value of $Fnamex$ is observed, then this implicitly partitions the values of $Afname$ into the observed value and the other values. Note: the probability $P(Fnamex|Afname \land Sex \land EFx = noerr)$ is independent of $Sex$.

**case 2:**　$EFx = sde$

$$P(Fnamex|Afname \land EFx = sde) =$$

$$\begin{cases} prsing(Fnamex) & \text{if } singlet(Fnamex, Afname) \\ 0 & otherwise \end{cases}$$

where, $singlet$ is a predicate to test whether variables $Fnamex$ and $Afname$ are a single letter apart or not. $prsing$ is a function to compute the probability for $EFx = sde$. For example, if $Fnamex = dave$ then $prsing(dave) = \frac{1}{100}$ (Note: 100 words can be generated by $Fnamex = dave$ which are a single letter apart from $Fnamex$ as each letter can be replaced by 25 possible letters). Note: again the probability $P(Fnamex|Afname \land Sex \land EFx = sde)$ is independent of $Sex$.

---

[4]http://www.census.gov/genealogy/names/

[5]The data available from U.S. Census Bureau is very noisy and incomplete to apply any of the zero frequency estimation approaches.

**case 3:** $EFx = ce$.

$P(Fnamex|Afname \wedge Sex \wedge EFx = ce) =$

$$\begin{cases} P(Fnamex|Sex = male) & \text{if } Sex = male \\ P(Fnamex|Sex = female) & \text{if } Sex = female \end{cases}$$

To compute the probability $P(Fnamex|Sex = male)$ and $P(Fnamex|Sex = female)$ we use the male name file and female name file respectively. The predicate *intable* $(Fnamex, male)$ tests whether $Fnamex$ is in the male name file or not. If $Fnamex$ is in the male name file then function $lookup(Fnamex, male)$ computes the probability $P(Fnamex|Sex = male)$ by looking in the male name file. If Fnamex is not in the male name file then we consider $P(Fnamex|Sex = male)$ as the probability of a new name, $Pnew$, a very small probability.

The *if-then-else* structure can also be seen as a *decision tree* [Quinlan, 1986]. These representations have been used to represent context specific independence [Boutilier et al., 1996]. Generally speaking, the proposed representation generalizes the idea of context specific independence, because contexts are not only given by expression such as $variable_i = value$ but also by the expression such as $foo(variable_i, variable_j) = yes$. The decision tree representation of conditional probability table $P(Fnamex|Afname \wedge Sex \wedge EFx)$ is shown in Figure 2.

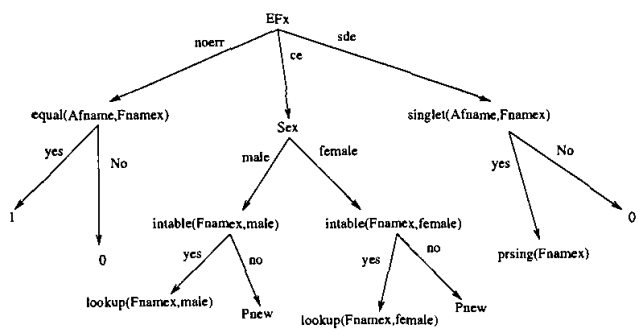

Figure 2: A decision tree representation of the CPT

In Figure 2 values of the leaves represent the probability for any world where all the variables in the path from the root to that leaf have corresponding values. For example, for the trees in Figure 2 the probability is $prsing(Fnamex)$ in any world when $EFx = sde$ and $singlet(Afname, Fnamex) = true$.

## 4 Large Domain Variable Elimination

The task of probabilistic inference is: given a Bayesian network with tree structured CPTs and evidence $E$, answer some probabilistic query, $P(X|E = e)$ i.e., the probability distribution over the random variable or variables $X$ given evidence $E = e$.

The inference algorithm for BN containing large variables is based on variable elimination, VE [Zhang and Poole, 1996]. In VE, a factor is the unit of data used during computation. A factor is a function over a set of variables. The factors can be represented as tables, where each row of the table corresponds to a specific instantiation of the factor variables. In VE the initial factors are conditional probability table. The main operations in this algorithm are:

- conditioning on observations
- multiplying factors
- summing out a variable from a factor

In large-domain VE, we represent the factors as decision trees, as shown in Figure 2.

Initially, the factors represent the conditional probability tables. For the intermediate factors that are created by adding and multiplying factors, we need to find the partitions of large variables dynamically for each assignment of small variables and partitions on other large variables.

### 4.1 Operations on Trees

In this section, we briefly describe two operations on which we build the operations: multiplying factors, and summing out variables from a factor.

**Tree Pruning (simplification)**

Tree pruning is used to remove redundant interior nodes or redundant subtrees of the interior nodes of a tree. We prune branches that are incompatible with the ancestors in the tree. In the simplest case, where we just have equality, we prune any branch where an ancestor gives a variable a different value. Where there are explicit sets, we can carry out an intersection to determine the effective constraints. We can then prune any branch where the effective constraint is that a variable is a member of the empty set. For example, if an ancestor specifies $X \in \{1, 2\}$ and a decendent specifies $X \in \{3\}$, the decendent can be pruned. Similarly for the "else" case, we can do set difference to determine the effective constraints. An example is shown in Figure 3. The tree on the left contains multiple interior nodes labelled $X$ along a single branch. The tree can be simplified to produce a new tree in which the subtree of the subsequent occurrence of $X$ which are not feasible are removed.

The correctness of the algorithm does not depend on whether we do complete pruning. We don't consider checking for compatibility of intensional representations (which may require some theorem proving); whether the algorithm can be more efficient with such operations is still an open question.

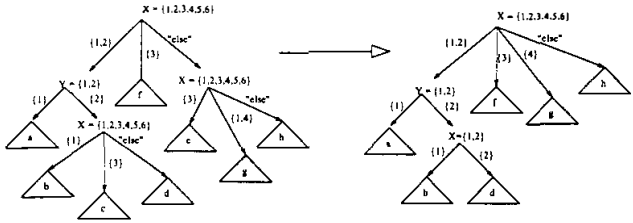

Figure 3: A tree simplified by removal of redundant subtrees (triangle denote subtrees)

**Merging Trees**

In VE, we need to multiply factors and sum out a variable from a factor. Both of these operations are built upon the merging trees operation.

Two trees $T1$ and $T2$ can be merged using operation $Op$ to form a single tree that makes all the distinctions made in any of $T1$ and $T2$, and with $Op$ applied to the leaves. When we merge $T1$ and $T2$, we replace the leaves of tree $T1$ by the structure of tree $T2$. The new leaves of the merged tree are labelled with the function, $Op$, of the label of the leaf in $T1$ and the label of the leaf in $T2$. We write $merge2(T1, T2, Op)$ to denote the resulting tree. If the labels of the leaves are constant, the leaf value of the new merged tree can be evaluated while merging the trees. If the leaf labels are *intensional* functions, one of the choices is when to evaluate the *intensional* function. When to evaluate the intentional functions can be considered as a secondary optimization problem. We always apply the pruning operation to the merged tree.

For example, Figure 4 shows tree $T2$ being merged to tree $T1$ with the addition (+) operator being applied. When we merge two trees and the $Op$ is a multiplication function then if the value at any leaf of $T1$ is zero, we keep that leaf of $T1$ unchanged in the merged tree. We do not put the structure of $T2$ at that leaf (as shown in Figure 5).

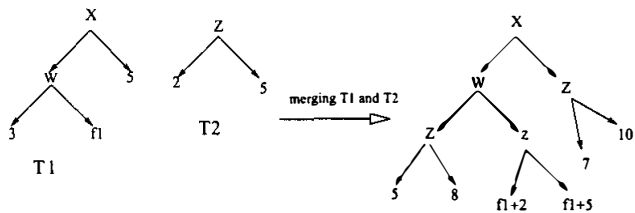

Figure 4: Merging tree $T1$ and $T2$ and leaf labels are combined using the *plus* function $merge2(T1, T2, +)$

We can extend the $merge2$ operator to a set of trees. We can define $merge(Ts, Op)$ where, $Ts$ is a set of trees and $Op$ is an operator, as follows. We choose a total order of the set, and carry out the following recursive procedure:

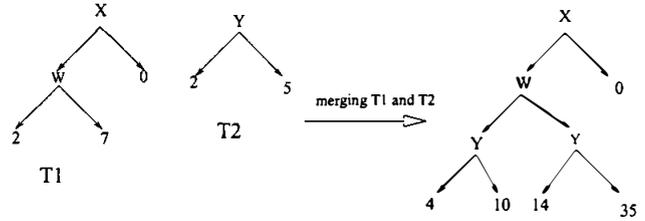

Figure 5: Merging tree $T1$ and $T2$ and leaf labels are combined using the *multiplication* function $merge2(T1, T2, \times)$

$merge(\{T_0, \ldots, T_n\}, Op)$
$= T_0$ if $n = 0$
$= merge2(merge(\{T_0, \ldots, T_{n-1}\}, Op), T_n, Op)$ if $n > 0$

### 4.2 Conditioning on Observations

When we observe the values taken by certain variables, we need to incorporate the observation into the factors. If a node is split on the values of the observed variable, the observed value of a variable is incorporated in the tree representation by replacing that node by its subtree that corresponds to the observed value. If a node split on an *intensional* function of the observed variable, the observed value of a variable is incorporated by replacing the occurrence of the variable by its observed value.

For example, when we observe $Fnamex = david$, then factor $f(EFx, Fnamex = david, Afname)$ becomes a function of $EFx$, and $Afname$. The tree representation of the new factor $f(EFx, Afname)$ is shown in Figure 6.

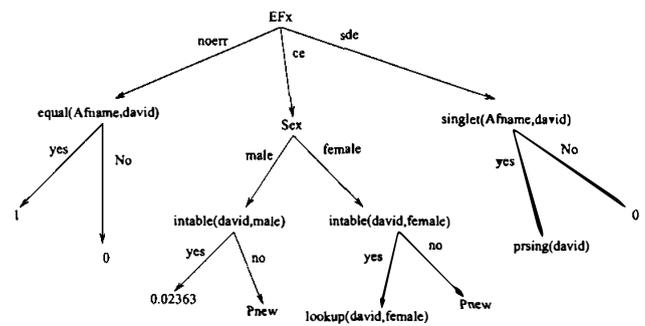

Figure 6: A Tree structured representation of new factor $f(EFx, Afname)$, i.e., the factor $f(EFx, Fnamex, Afname)$ after conditioning on $Fnamex = david$

In Figure 6, the predicate *equal* gives us the possible value for $Afname$ which is equal to $david$. That is, in the context of $EFx = noerr$, we are implicitly partitioning $Afname$ into $\{david\}$ and all of the other names. Similarly, for $EFx = sde$, we are implicitly partitioning the values of $Afname$ into those names which are a single letter apart from $david$, and all of the other names.

The computation of predicates *equal* and *singlet* is delayed until we sum out the variable $Afname$. We can now compute the predicate $intable(david, male)$ and $intable(david, female)$ to simplify the tree after conditioning on observation $Fnamex = david$. As $david$ appears in the male name file, the subtree at node $intable(david, male)$ is replaced by the value of $lookup(david, male)$ which is 0.02363. As $david$ doesn't appear in the female name file, the subtree at node $intable(david, female)$ is replaced by the probability of new name, $Pnew$.

### 4.3 Multiplication of Factors

In variable elimination, to eliminate $Y$, we multiply all of the factors that contain $Y$, then sum out $Y$ from the resulting factor. In this section we describe how to multiply factors represented as trees.

Suppose $T$ is the set of trees that represent the factors that involve $Y$. We need to form the product $merge(T, \times)$, from which we will sum $Y$. We always apply the pruning operation to the resulting tree.

For example, suppose that we have observed $Fnamex = david$ and $Fnamey = davig$ and we want to eliminate the variable $Afname$ from Figure 1. To eliminate variable $Afname$ we need to multiply all the factors that contain variable $Afname$. The factors $f1(Fnamex = david, EFx, Sex, Afname)$, $f2(Fnamey = davig, EFy, Sex, Afname)$, and $f3(Afname, Sex)$ contain variable $Afname$. As shown in Figure 7, $T1$, $T2$ and $T3$ are the decision tree representation of $f1$, $f2$, and $f3$ respectively. After multiplying factors $f1$, $f2$, and $f3$ we get a new factor $f(EFx, EFy, Sex, Afname)$ of variables $Efx$, $EFy$, $Sex$, and $Afname$. Part of the tree representation, $T$, of the new factor, $f$, is shown in Figure 7.

### 4.4 Summing Out Variable $Y$

Suppose $T$ is the tree representation for the factor resulting from multiplying all trees that contain variable $Y$. Now, we need to sum out the variable $Y$ from $T$ in order to get the tree representation, $T'$, of the new factor.

In large domain VE, summing out a variable is complicated because we can have *intensional* functions at the nodes as well as on the leaves of the tree. To sum out a variable $Y$ from tree $T$, at each leaf we need to compute the *probability mass* for all the values of $Y$ that end up at each leaf.

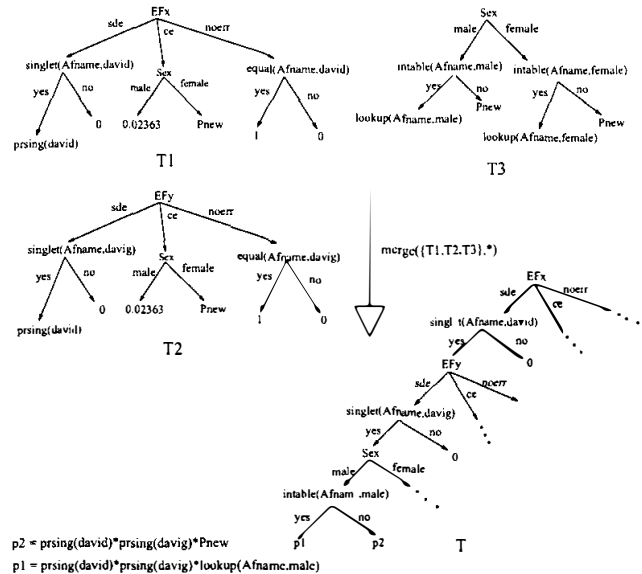

Figure 7: A decision tree representation, $T$, of new factor $f$, after multiplying factors $f1$, $f2$, and $f3$ (* represents multiplication operator)

If the label at the leaf is a constant, the *probability mass* of a leaf is the product of the label and the number of values of $Y$ which satisfies all the predicates from the root to this leaf. If the label at the leaf is a function, the *probability mass* of a leaf is computed by summing the value of the leaf for each value of $Y$ that satisfies all the predicates from the root to this leaf. How to evaluate this depends on the actual function.

Once we have the *probability mass* at each leaf, we need to sum the subtrees that correspond to different blocks (subsets) for a partition of $Y$. We need to do this for every context (i.e., for every assignment of ancestors).

These two steps, for computing $T'$ from $T$, are combined in the algorithm shown in Figure 8. We traverse the tree $T$ in a top-down manner. At each internal node, we determine if the test for the split depends on the summing variable. If so, we sum out $Y$ from each subtree recursively and then merge them together using plus opeartor. If not, we recursivly call each subtree. In order to determine the probability mass at the leaves, we keep track of all the predicates that refer to $Y$ during the recursion. If a node is a leaf, we compute the *probability mass* of the leaf for all the values of $Y$, which satisfy all the predicates from the root to this leaf.

Note: When we sum out a small variable and the nodes in the tree split on the values of the small summing variable, the algorithm shown in Figure 8 is simple because in this case the *probability mass* of a leaf is the same as the label on the leaf.

As an example, suppose we want to sum out the variable $Afname$ from factor $f(EFx, EFy, Sex, Afname)$

Function Sum (T, Y, Context) **returns** a decision tree **T'**
Input : $T$, the root of the decision tree, $Y$, the summing variable
initially Context is true
  **if** $T$ is an internal node **then**
    $fun \leftarrow$ function at which $T$ is tested
    **if** $Y \in fun$ **then**
      $T_0, \ldots, T_n \leftarrow$ subtrees of $T$
      $C_0, \ldots, C_n \leftarrow$ values of $fun$ for $T_0, \ldots, T_n$
      $T'_i \leftarrow Sum\,(T_i, Y, Context \wedge (fun = C_i))$
      $TT \leftarrow merge\,(\{T'_0, \ldots, T'_n\}, +)$
      return $TT$
    **else**
      $T' \leftarrow$ a new node with test on $fun$
      $T_0, \ldots, T_n \leftarrow$ subtrees of $T$
      $T'_i \leftarrow Sum\,(T_i, Y, Context)$
      add $T'_0, \ldots, T'_n$ to $T'$
      return $T'$
    **end if**
  **else if** $T$ is leafnode **then**
    let $p \leftarrow$ leaf label
    $p' \leftarrow \sum_{\forall y \in dom(Y), Context=true} p$
    leaf label $\leftarrow p'$
    Return $T$
  **end if**

Figure 8: Algorithm for computing decison tree $T'$ after summing out variable $Y$ from the decison tree $T$

as computed in Section 4.3. The tree representation $T$ of $f$ is shown in Figure 7. After we sum out the variable $Afname$ from $f$ we get a new factor $f'$ ($EFx, EFy, Sex$) of variables $EFx$, $EFy$, and $Sex$. The tree representation $T'$ of new factor $f'$ is shown in Figure 9.

In the next section we show how *probability masses* $p1'$ and $p2'$ can be computed efficiently without actually enumerating the values of $Afname$.

### 4.4.1 Evaluation of $p1'$ and $p2'$

Let us first consider the computation of the probability mass, $p1'$.

$$p1' = \sum_{\forall Afname=afname \in dom(Afname)(C1 \wedge C2 \wedge C3=true)} p1$$

where, $C1 = (singlet(Afname, david) = yes)$, $C2 = (singlet(Afname, davig) = yes)$, and $C3 = (intable(Afname, male) = yes)$

As shown in Figure 9, $p1$ is a function of $Afname$, to compute the value of $p1'$ we need to compute $p1$ for all values of $Afname$ that satisfy the predicates $C1$, $C2$, and $C3$. That is, those values of $Afname$ which exist in the male name file and single letter apart from both $Afname = david$

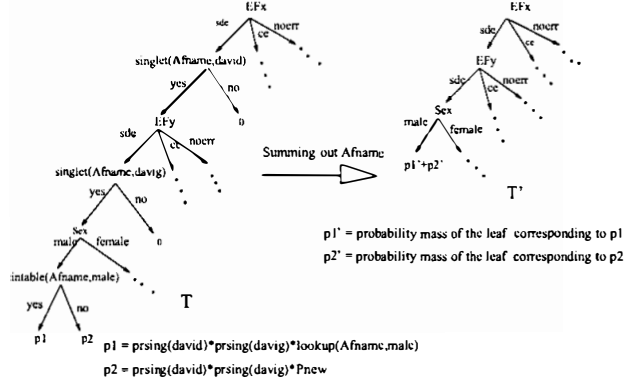

Figure 9: A decision tree representation, $T'$, of new factor $f'$ after summing out the variable $Afname$ from factor $f$ (∗ represents multiplication operator)

and $Afname = davig$. We can compute these values of $Afname$ without actually enumerating the values of $Afname$ by using an efficient data structure for representing the male name file and the female name file. Now, we can query to the male name file representation to get the values of $Afname$ that are single letter apart from both $david$ and $davig$, we get $Afname = \{davis\}$. Thus,

$$p1' = \sum_{Afname=\{davis\}} p1$$
$$= prsing(david) \times prsing(davig) \times Pdavis$$

where, $Pdavis$ is the probability of name $davis$ from male name file

Let us now consider the computation of *probability mass*, $p2'$.

$$p2' = \sum_{\forall Afname=afname \in dom(Afname)(C1 \wedge C2 \wedge C4=true)} p2$$

where, $C4 = (intable(Afname, male) = no)$

As shown in Figure 9, $p2$ is a not a function of $Afname$, to compute the value of $p2'$ we don't need the values of $Afname$ that satisfy the predicates $C1$, $C2$, and $C4$. But, we need the count of the values of $Afame$.

To count efficiently the number of values of $Afname$ that are single letter apart from both $david$ and $davig$, we first generate the patterns of names that are a single letter apart from $david$. For example, $?avid$, where ? is any letter except $d$. After generating these patterns we test which of these patterns makes the predicate $singlet(Afname, davig) = yes$. Here, the pattern $davi?$ makes the predicate $yes$ if $? \neq d \wedge ? \neq g$. Thus, the possible number of values for $Afname$ is 24 that are a single letter apart from both $david$ and $davig$[6]. Out of these 24 values of $Afname$ we have already found that one value exist in male name file (during the computation of $p1'$). Thus, there are only 23 values of $Afname$ that satisfy $C1 \wedge C2 \wedge C4$. Thus,

$$p2' = 23 \times prsing(david) \times prsing(davig) \times Pnew$$

---

[6]As there are 26 letters.

### 4.5 Computing Posterior

To compute the posterior we first condition on the observed variables and then sum out all non-observed, non-query variables one by one. We can compute the posterior by multiplying the remaining factors and normalizing the remaining factor.

When we query a large variable, we would typically return an intensional representation of the distribution, which we can use to answer queries about the distribution.

## 5 Conclusion

In this paper we present an inference algorithm for a belief network that contains random variables with large or even unbounded domains. Our inference algorithm, *large domain variable elimination*, is based on the variable elimination algorithm. The main idea is to partition the domain of a large variable in equivalence classes for which we have the same conditional probability for particular values (or particular subset) of other random variables. We construct these subsets dynamically during the inference. These equivalence classes can be described extensionally and intensionally. Intensional representation allows us to compute the query in terms of parameters and then the answer to specific queries are computed by plugging the values of the parameters.

## Acknowledgements

This work was supported by NSERC Research Grant OGPOO44121 and The Institute for Robotics and Intelligent Systems. Thanks to Valerie McRae for providing the useful comments.

## References

Bell, G. B. and Sethi, A. [2001]. Matching records in a national medical patient index, *Communication of the ACM*, Vol. 44.

Boutilier, C., Froedman, N., Goldszmidt, M. and Koller, D. [1996]. Context-specific independence in Bayesian networks, *In Proceeding of Thirteenth Conf. on Uncertainity in Artificial Intelligence (UAI-96)*, pp. 115–123.

Chen, S. F. and Goodman, J. T. [1998]. An empirical study of smoothing techniques for language modeling, *Technical Report TR-10-98, Computer Science Group, Harvard University*.

Cohen, J. D. [1997]. Recursive hashing functions for n-grams, *ACM Transactions on Information Systems*, Vol. 15(3), pp. 291–320.

Fellegi, I. and Sunter, A. [1969]. A theory for record linkage, *Journal of the American Statistical Association*, pp. 1183–1210.

Friedman, N. and Singer, Y. [1998]. Efficient Bayesian parameter estimation in large discrete domains, *Proceedings of Neural information Processing Systems*.

Geiger, D. and Heckerman, D. [1996]. Knowledge representation and inference in similarity networks and Bayesian multinets, *Journal of the Artificial Intelligence*, Vol. 82, pp. 45–74.

Gill, L. [1997]. Ox-link: The Oxford medical record linkage system, *Record Linkage Techniques*, National Academy Press, pp. 15–33.

Good, I. [1953]. The population frequencies of species and the estimation of population parameters, *Journal of the American Statistical Association*, pp. 237–264.

Odell, J., Violative, V. and Woodland, P. [1995]. A one pass decoder design for large vocabulary recognition, *Proceedings of the DARPA Human Language Technology Workshop*.

Pearl, J. [1988]. *Probabilistic reasoning in intelligent systems: networks of plausible inference*, Morgan Kaufmann Publishers Inc.

Quinlan, J. [1986]. Induction of decision trees, *Machine Learning*, Vol. 1, pp. 81–106.

Zhang, N. L. and Poole, D. [1996]. Exploiting causal independence in Bayesian network inference, *Artificial Intelligence*, Vol. 5, pp. 301–328.